%% file: emnlp-ijcnlp-2019.tex
\documentclass[11pt,a4paper]{article}
\usepackage[hyperref]{emnlp-ijcnlp-2019}

\usepackage{tikz}
\usepackage{scalerel}

\usepackage{microtype}

\usepackage{times}
\usepackage{latexsym}
\usepackage{enumitem}
\usepackage{todonotes}
\usepackage{amsmath,amsfonts,amssymb}
\usepackage{subcaption}
\usepackage{multirow}
\usepackage{scrextend}
\usepackage{booktabs}
\usepackage{tabularx}
\usepackage{url}
\usepackage{graphicx}
\usepackage{multirow}
\usepackage{amsthm}
\usepackage{amssymb}
\usepackage{amsmath}
\usepackage{amsfonts}
\usepackage{xcolor}
\usepackage{mdwlist}
\usepackage{microtype}
\usepackage{tabularx}
\usepackage{booktabs}
\usepackage{subcaption}
\usepackage{arydshln} 

\usepackage{cleveref}
\crefname{section}{§}{§§}
\Crefname{section}{§}{§§}

\usepackage{url}

\aclfinalcopy 

\DeclareMathOperator*{\argmax}{arg\,max}

\newcommand{\tikztriangledown}[1][blue,fill=blue]{\tikz{\filldraw[draw=#1,fill=#1]  (0.1cm, 0) -- (0,0.2cm) --  (0.2cm,0.2cm)  ;}}

\newcommand{\tikztriangleup}[1][green,fill=green]{\tikz{\filldraw[draw=#1,fill=#1] (0,0) --
(0.2cm,0) -- (0.1cm,0.2cm);}}

\title{What do Deep Networks Like to Read?}

\author{\bf Jonas Pfeiffer$^{1}$\thanks{{ } Both authors contributed equally to this work.} , Aishwarya Kamath$^{2*}$, Iryna Gurevych$^1$, Sebastian Ruder$^{3,4}$\thanks{{ } Sebastian is now affiliated with DeepMind.}\\
$^1$ Technische Universit\"at Darmstadt\\
$^2$ New York University\\
$^3$ Insight Centre, NUI Galway  \\
$^4$ Aylien Ltd., Dublin \\
\texttt {\{pfeiffer,gurevych\}@ukp.informatik.tu-darmstadt.de} \\ 
\texttt {ask762@nyu.edu}
}

\date{}

\begin{document}
\maketitle
\begin{abstract}

Recent research towards understanding neural networks probes models in a top-down manner, but is only able to identify model tendencies that are known a priori. 
We propose  \textbf{S}usceptibility \textbf{I}dentification through \textbf{F}ine-\textbf{T}uning  (SIFT), a novel abstractive method that uncovers a model's preferences without imposing any prior. By fine-tuning an autoencoder with the gradients from a fixed classifier, we are able to extract propensities that characterize different kinds of classifiers in a bottom-up manner. We further leverage the SIFT architecture to rephrase sentences in order to predict the opposing class of the ground truth label, uncovering potential artifacts encoded in the fixed classification model. 
We evaluate our method on three diverse tasks with four different models. We contrast the propensities of the models as well as reproduce artifacts reported in the literature.


\end{abstract}

\section{Introduction}

Recent research on understanding and interpreting neural networks in natural language processing has progressed in two main directions: 1) Approaches to probe particular capabilities of models based on synthetic datasets
\cite{Adi2017,Conneau2018a,Zhu2018,Peters2018a}, e.g. if they capture information regarding the length of a sequence; 
and 2) approaches that extract or assign weight to \emph{rationales}, such as n-grams in the input that are indicative of the final prediction \cite{Lei2016,Ribeiro2016,Ribeiro2018,Murdoch2018a,Bao2018}. While such rationales can help motivate individual predictions, they fall short of uncovering a model's inherent preferences.

To fill this gap, we propose an \emph{abstractive} method for understanding neural networks applied to text in a \emph{bottom-up} fashion.  
Inspired by recent work on understanding convolutional neural networks in computer vision \cite{Palacio2018}, we propose  \textbf{S}usceptibility \textbf{I}dentification through \textbf{F}ine-\textbf{T}uning  (SIFT).
SIFT passes the output of an \textit{autoencoder} (AE) into a pretrained classifier with frozen weights. We fine-tune the AE with the gradients from the classifier using the Straight-Through Gumbel-Softmax estimator \cite{Jang2017}.
During fine-tuning, the AE learns to reformulate parts of the input that are irrelevant to the classifier and to only retain information that it deems useful. Inspecting the reconstructed samples thus gives us a window into what the classifier likes to read. 
In contrast to extractive approaches, SIFT is able to leverage information from large amounts of unlabelled data via pretraining, allowing us to make use of a broader vocabulary of words and knowledge for model introspection.

\paragraph{Contributions}
 We conduct a multi-pronged analysis of three popular sentence classification models---an LSTM-based text classifier, a CNN variant \cite{Kim2014} and a Deep Averaging Network \cite[DAN;][]{Iyyer2015}---to facilitate comparison between their preferences (\S \ref{sec:classifier-comparison}). We are able to extract patterns, which are correlated with the respective model architecture. In an attempt to extract propensities of the pre-trained classification models, we 
uncover terms and phrases whose presence in the input causes the classifier to predict a given class (\S \ref{sec:opposite-labels}). Besides the sentence classification tasks, we additionally report results for a two-sentence setup on Natural Language Inference with the model of \citet{Bowman2015}.
In all cases we implement simple models and focus on uncovering fundamental dependencies instead of trying to disentangle the various moving parts in more complex models.

\section{Related work}

\paragraph{Understanding neural networks} Most recent research on understanding neural networks utilizes \emph{challenge sets}, test suites that seek to evaluate particular properties of a model; see \cite{Belinkov2019} for an overview. Among these, \citet{Adi2017} investigate if different sentence representations can encode sequence length, word content, and order, while \citet{Conneau2018a} test for simple syntactic properties such as constituency tree depth, tense, and subject number.
\citet{Zhu2018} generate triplets of sentences to explore how changes in the syntactic structure or semantics affect the similarities between the embeddings.
\citet{Peters2018a} use part-of-speech tagging and constituency parsing for probing contextual representations at different layers.
The drawback of these challenge sets is that they only allow for inspecting characteristics that have to be defined a priori. 
A contrasting approach is to investigate the behaviour of individual neurons \cite{LiCHJ16, Bau2019}. This process, however, can quickly become cumbersome as the role of individual neurons differs between models and only yields local insights. 
In contrast, our model produces reformulations at a global level for the target model making it easier to inspect. Our model is inspired by the work of \citet{Palacio2018} who also fine-tune an AE with a fixed classifier. Their model however, is only able to deal with continuous inputs and outputs. On the other had, our model learns to reconstruct discrete sequences.

\paragraph{Interpreting model predictions} Much work on interpreting model predictions focuses on extracting \emph{rationales}---subsets of words from the input that are short, coherent, and suffice to produce a prediction.
\citet{Lei2016} jointly train a generator with the model and extract rationales by forcing the model's prediction based on the rationale to be close to the model's prediction on the original input. \citet{Ribeiro2016} propose LIME, which approximates a model locally with a sparse linear model, focusing on keywords that are strongly associated with a class. \citet{Ribeiro2018} propose Anchors, high-precision rules that represent local, 'sufficient' conditions for predictions and an algorithm to compute them for black-box models. \citet{Murdoch2018a} proposes contextual decomposition, a method to decompose the output of LSTMs and identify words and phrases that are associated with different classes.
\citet{Bao2018} use annotated rationales as supervision for attention.
In contrast to these approaches, our method is not limited to extracting words or phrases from the input, but learns to paraphrase and condense relevant information. 
Of these extractive methods, the methods by \citet{Lei2016} and \citet{BastingsAT19} are most similar to ours as they also train a generator in tandem with a model. 
Our approach can also be seen as a way to elicit desired behaviour from an algorithm, similar to \citet{Buck2018} who learn to reformulate questions. While their method is restricted to question answering, our framework is potentially applicable to any arbitrary NLP task.

\paragraph{Data set artifacts} 

Recent work has shown that data set artifacts, which are introduced as a by-product of crowd-sourced annotations, leak information about the target label \cite{Gururangan2018, poliak2018hypothesis, tsuchiya2018performance}. Machine learning algorithms are able to exploit these artifacts to predict the correct class without actually solving the task at hand \cite{sakaguchi2019winogrande}. Recent works mostly apply adversarial filtering approaches \cite{zellers2018swag, sakaguchi2019winogrande} to reduce the consequences of the aforementioned bias but have not focused on \emph{identifying} the artifacts that have been encoded by the model, which we investigate in this work.


\section{SIFT}

Our proposed \textbf{S}usceptibility \textbf{I}dentification through \textbf{F}ine-\textbf{T}uning (SIFT) framework can be used to analyze \textit{any} sentence-level pre-trained model. It consists of two stages: \textit{Pretraining} and \textit{Fine-tuning}. 

\paragraph{Pretraining} \label{sec:pretraining}


We define our autoencoder as an LSTM-based sequence-to-sequence model \cite{Sutskever2014} augmented with attention \cite{Bahdanau2015} that is trained to reliably reconstruct the 1B Word Benchmark \cite{chelba2013one} and the IMDb movie review dataset \cite{Maas2011}. We initialize the word embeddings of all models (both AEs and classifiers) with the top 30k 100-dimensional GloVe embeddings \cite{Pennington2014}. The encoder (bidirectional, hidden size $512$) and decoder (uni-directional, hidden size $1024$) of the AE both have one layer. 
The encoder's last hidden states are maxpooled to initialize the decoder. 
Prior to fine-tuning, we make a preliminary run of the trained AE on data from the respective classification task in order to adapt it.


%

\paragraph{Fine-tuning the AE} \label{sec:fine-tuning}

To achieve our objective of deducing preferences of the classifier, 
we fine-tune the decoder of our pre-trained AE using gradients from the pre-trained classifier. We hypothesize that this would have the effect of bring the text produced by the decoder into a form that makes it more amenable to the specific classifier, thus revealing its preferences and idiosyncrasies. 

In order to update the parameters of the decoder, we must propagate the gradient through the non-differentiable operation of sampling from a categorical distribution. To overcome this, we employ the Straight-Through Gumbel-Softmax estimator \cite{Jang2017} defined as:
\begin{equation}
\tilde{y}_j = \frac{\exp((\log(z_j) + g_j)/\tau)}{\sum^V_{v=1} \exp((\log(z_v) + g_v)/\tau)}
\end{equation}


\begin{figure}[!t]
    \begin{subfigure}{\linewidth}
      \centering
         \includegraphics[width=\linewidth]{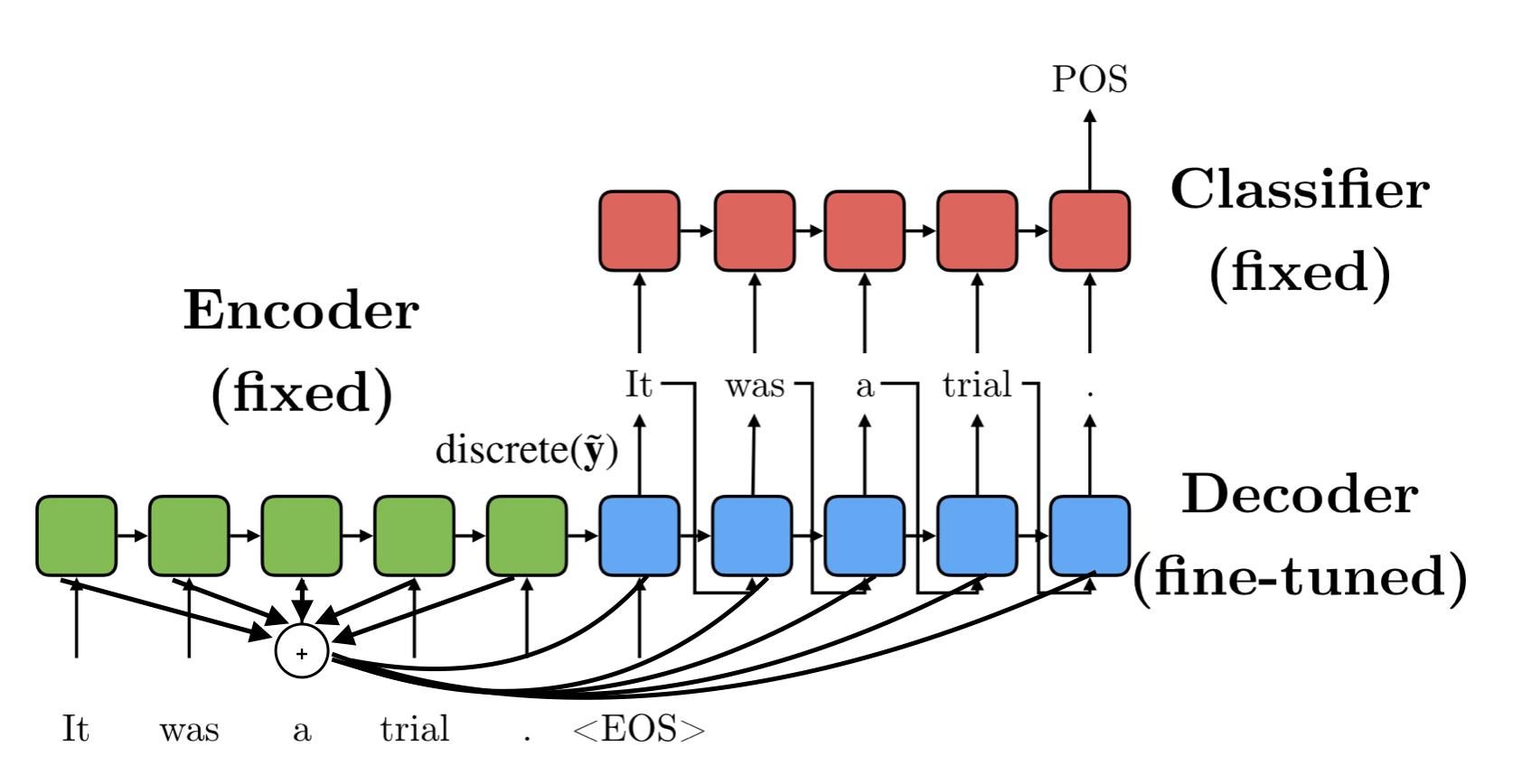}
    \caption{Autoencoder fine-tuning (forward pass)} \label{fig:fine-tuning_forward}
    \end{subfigure}
    
    \begin{subfigure}{\linewidth}
      \centering
         \includegraphics[width=\linewidth]{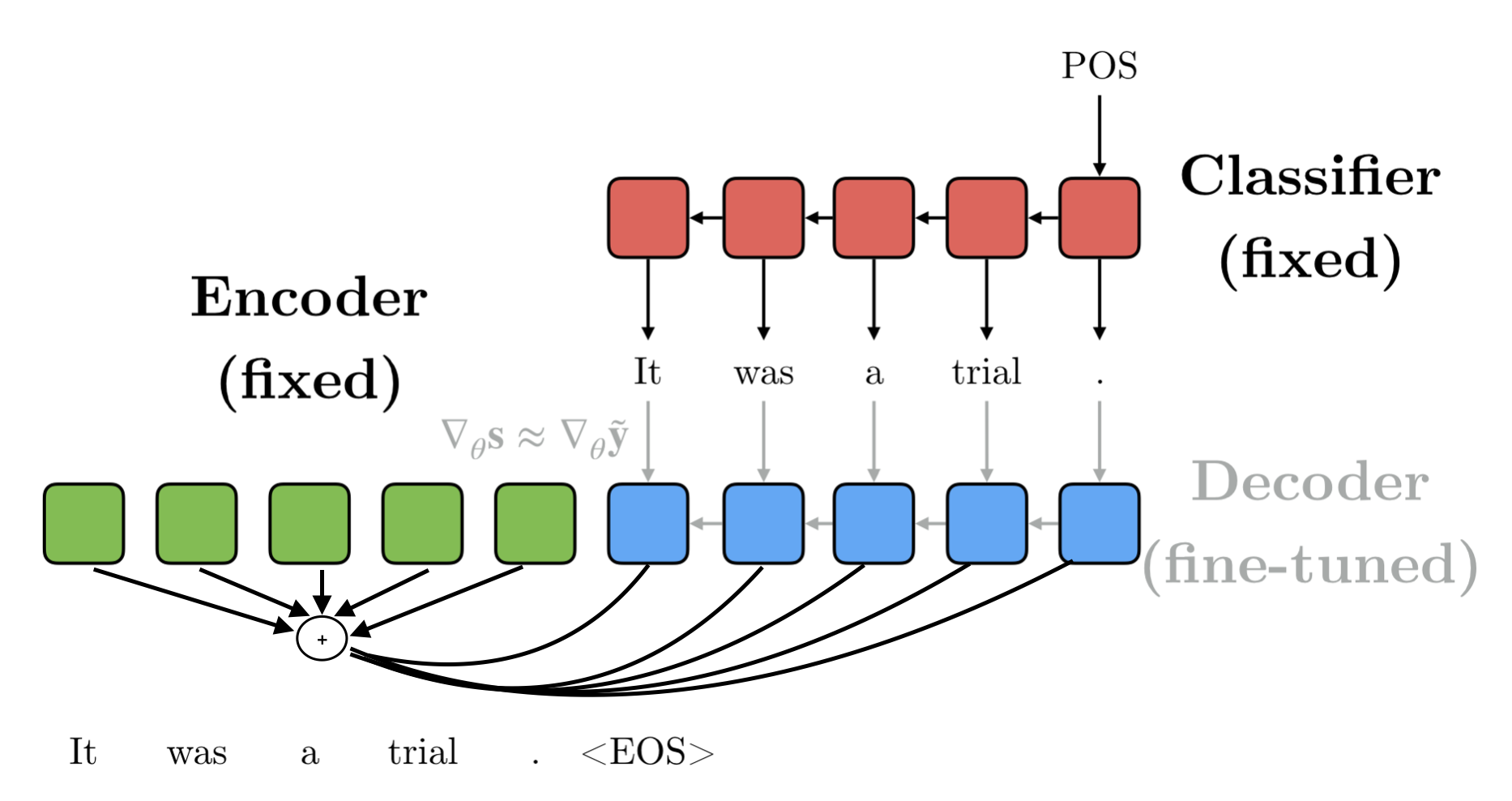}
    \caption{Autoencoder fine-tuning (backward pass)} \label{fig:fine-tuning_backward}
    \end{subfigure}
    \caption[]{Fine-tuning the decoder of the AE with the gradients of the fixed classifier. During the forward pass (a), the Gumbel-Softmax sample of the AE is discretized
    as input for the classifier. During the backward pass (b), gradients are back-propagated through the classifier, through the generated output---using the gradient of the Gumbel-Softmax estimator---to the decoder. Arrows indicate how the gradient is propagated; only parameters with gray arrows are updated.}
\label{fig:AE_fine-tuning}
\vspace{-4mm}
\end{figure}

\noindent where $j$ is the target position of vocabulary $V$, $z$ are logits from the final layer, $\tau$ is a temperature parameter, and  $\mathbf{g} \in \mathbb{R}^{|V|}$ corresponds to samples from the Gumbel distribution $g_v \sim - \log(- \log(u_v))$
with $u_i \sim \mathcal{U}(0, 1)$ being the uniform distribution. As $\tau \rightarrow 0$, the softmax becomes an $\argmax$ and the Gumbel-Softmax distribution approximates more closely the categorical distribution.


During the forward pass as shown in Figure \ref{fig:fine-tuning_forward}, we discretize this continuous sample using $\mathrm{one\_hot}(\argmax(\tilde{\mathbf{y}}))$
which is then used to lookup the corresponding word embedding to be passed forward to the classifier.
During the backward pass depicted in Figure \ref{fig:fine-tuning_backward}, we approximate the gradient of the discrete sample $\nabla_\theta \mathbf{s}$ with the gradient of our continuous approximation $\nabla_\theta \tilde{\mathbf{y}}$.
We found the fine-tuning of the decoder to be extremely sensitive to the choice of hyper-parameters and find that a learning rate of $5e^{-5}$  and Gumbel-Softmax temperature of $0.9$ work well for most tasks. We found that including word dropout \cite{Iyyer2015} on the output of the decoder greatly improved stability of training.

\input{05-analysis.tex}
\section{Conclusion}



We have proposed a bottom-up approach for extracting a model's preferences and consequently understanding opaque neural network architectures better. We have sought to overcome the difficulty of evaluating an abstractive unsupervised approach by means of a multi-pronged analysis, highlighting how our approach can be used to analyze examples, compare classifiers, and generate sentences with opposite labels. We were able to uncover significant differences of what the respective architectures want to read by \textit{sifting} out what models have encoded in order to understand their limits and create fair representations in the future.  
We believe that our approach is an important first step towards abstractive interpretability methods and bottom-up bias identification in NLP.

\bibliography{emnlp-ijcnlp-2019}
\bibliographystyle{acl_natbib}
\newpage
\appendix
\setcounter{table}{0}
\renewcommand{\thetable}{A\arabic{table}}

\section{Supplemental Material}
\label{sec:supplemental}

\begin{figure}[!htp] 
    \centering
    \includegraphics[width=\linewidth]{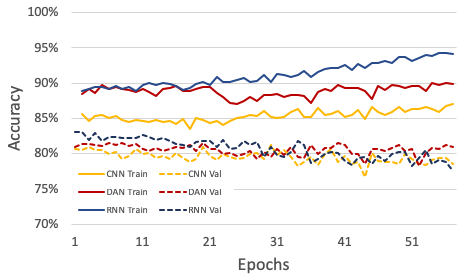}
    \caption{Train and validation accuracy of SIFT in the Classifier Inspection setting for CNN, RNN, and DAN models on SST-2.}
    \label{fig:Train_val_acc_sent}
\end{figure}

\begin{table}[htp]
\centering{\footnotesize
\begin{tabular}{llll}
\toprule
                           & \textbf{RNN}  & \textbf{CNN}  & \textbf{DAN}  \\ \toprule
Nouns                      & \tikz\draw[gray,fill=gray] (0,0) circle (.7ex);  $-1$   & \tikz\draw[gray,fill=gray] (0,0) circle (.7ex);  $+3$ & \tikztriangledown    ~ $-\textbf{9}$   \\ 
DT                         & \tikztriangleup  ~ $+\textbf{23}$ & \tikztriangledown ~ $-15$ & \tikz\draw[gray,fill=gray] (0,0) circle (.7ex); ~ $+\textbf{2}$ \\ 
Verbs                      & \tikztriangledown ~ $-\textbf{14}$ & \tikztriangledown ~ $-13$ & \tikztriangledown ~ $-6$  \\ 
Adj.                       & \tikztriangledown ~  $-12$ & \tikztriangledown ~ $-\textbf{23}$  & \tikz\draw[gray,fill=gray] (0,0) circle (.7ex); ~ $-1$ \\ 
Prep.                      & \tikz\draw[gray,fill=gray] (0,0) circle (.7ex); ~ $+5$ & \tikztriangledown ~ $-\textbf{16}$ & \tikz\draw[gray,fill=gray] (0,0) circle (.7ex); ~ $+3$ \\ 
Punct.                     & \tikztriangledown ~ $-18$ & \tikztriangleup ~ $+\textbf{78}$ & \tikz\draw[gray,fill=gray] (0,0) circle (.7ex); ~ $0$ \\ 
\textless{}U\textgreater{} & \tikztriangleup ~ $+52$   & \tikztriangleup $+\textbf{68}$ & \tikztriangledown ~ $-29$ \\
\bottomrule
\end{tabular}\caption{Part-of-Speech (POS) changes in PubMed: \tikztriangleup, \tikztriangledown , and \tikz\draw[gray,fill=gray] (0,0) circle (.7ex);  indicate that the number of occurrences have increased, decreased or stayed the same through fine-tuning respectively. The numbers indicate the changes in percentage points wrt to the original sentece. A score of 0 would thus mean that fine-tuning has not changed the number of words. }
\label{table:A1}}
\end{table}

\begin{table}[]
\centering{\footnotesize
\begin{tabular}{ c  ll }
\toprule
\multicolumn{1}{l}{}  & \multicolumn{2}{c}{\textbf{SNLI}} \\
\multicolumn{1}{l}{}    & Neutral       & Contradiction      \\ \cmidrule(ll){2-3}
\multirow{5}{*}{\textbf{PMI}}        & winning           & nobody           \\
                         & vacation           & cats           \\
                       & favorite           & sleeping           \\
                          & sad           & cat           \\
                      & owner           & tv           \\ \cmidrule(ll){2-3}
\multirow{5}{*}{\textbf{SIFT}}        & one         &  sitting            \\
                         & out           &    cave        \\
                        & batman           & women           \\
                      & on           & hand           \\
                  & with          & one         

                      \\ 
                        \hdashline
Accuracy &         $\textit{30\%}$          & $\textit{ 70\%}$

                      \\ \hline
 \textit{Corr}          & $\textit{-0.0059}$          & $\textit{ 0.366}$        
                      \\                
                      \bottomrule
\end{tabular}}
\caption{Top 5 PMI and SIFT terms for SNLI in the Artifact Detection setting. The second last row depicts the SIFT accuracy on the test set for the one class, flipped label setting. The last row indicates the correlation of the PMI and SIFT list using weighted Kendall's tau correlation \cite{Shieh1998}. Note that this setting is very difficult for the current SIFT architecture as the decoders do not have access to the information of the paired sentence. While \textit{Entailment} to \textit{Contradiction} works well reaching $70\%$ accuracy, \textit{Contradiction} to \textit{Neutral} is only able to fool the classifier in $30\%$ of the cases. We hypothesize that the good results for \textit{Entailment} to \textit{Contradiction} result from the fact that for a sentence pair $A$ and $B$, if we find a reformulation  $A'$ such that $A'$ contradicts $A$, $A'$ will most likely also contradict $B$. Thus the decoder simply needs to generate a sentence which contradicts its input.  }
\label{table:A2}
\end{table}


\begin{table*}[htp]
\centering

\begin{tabular}{ll}
\toprule
O P  & adults with \textbf{their kids} are riding on a small \textbf{red} train .                                 \\
G P & \begin{tabular}[c]{@{}l@{}}adults with \textbf{people} are riding small are train on on \textbf{sleeping} on on  \end{tabular}                              \\ \hline
H  & \begin{tabular}[c]{@{}l@{}}there are people on the train .\end{tabular}                  \\ \toprule

O P  & \begin{tabular}[c]{@{}l@{}}a black and white \textbf{dog} running through shallow water . \end{tabular}  \\
G P & \begin{tabular}[c]{@{}l@{}}with black white \textbf{cat} it water a small bed a on surface water  \end{tabular}    \\ \hline
H  & \begin{tabular}[c]{@{}l@{}} two dogs running through water . \end{tabular}                       \\ \toprule

O P  & \begin{tabular}[c]{@{}l@{}}two \textbf{boys} in green and white uniforms play \textbf{basketball} with two \textbf{boys} in blue and white \\ uniforms .  \end{tabular}  \\
G P & \begin{tabular}[c]{@{}l@{}}with \textbf{girls} in white uniforms . and girls black white \textbf{women} in \textbf{women} black \textbf{cats} black \textbf{sitting} \\ two   \end{tabular}    \\ \hline
H  & \begin{tabular}[c]{@{}l@{}} two different teams are playing basketball .  \end{tabular}                       \\ \toprule

O P  & \begin{tabular}[c]{@{}l@{}}very large group of old people riding in boats down a river . \end{tabular}  \\
G P & \begin{tabular}[c]{@{}l@{}}the \textbf{small} large group old riding boats the in in \textbf{sitting} a on on  \end{tabular}    \\ \hline
H  & \begin{tabular}[c]{@{}l@{}}mob of elderly riding water crafts . \end{tabular}                       \\ \toprule

\end{tabular}
\caption{Examples of premise sentences generated when controlled to produce a \textit{contradiction} from and \textit{entailment} with O standing for Original, G for generated H for Hypothesis and P for Premise. In this setting we only fine-tune the Premise as to not confuse SIFT by moving two independent parts. We highlight the terms we find most likely have fooled the classifier. Although many sentence pairs do not actually contradict each other, the classifier labels it as such, indicating that it has fixated on artifacts i.e. `sleeping', `cats', `sitting'}
\label{table:Flipped_examples}
\end{table*}

\renewcommand{\arraystretch}{1.1}
\begin{table*}[htp]
\centering
\begin{tabular}{ll}
\toprule

Orig & \begin{tabular}[c]{@{}l@{}} it 's a great deal of \textless{}u\textgreater{} and \textbf{ very little} steak .\end{tabular} \\ \hline
DAN  & \begin{tabular}[c]{@{}l@{}}it 's 's great yet of \textless{}u\textgreater{} and \textbf{very good} makes while \end{tabular}     \\ \hdashline
CNN  & \begin{tabular}[c]{@{}l@{}}it 's a great it of it and very good it . \textless{}u\textgreater{} . a . makes , so \end{tabular}                  \\ \hdashline
RNN  & \begin{tabular}[c]{@{}l@{}}it 's a great deal of \textless{}u\textgreater{} and very good sweet .\end{tabular}                        \\ \toprule

Orig & \begin{tabular}[c]{@{}l@{}} \textbf{fails} to bring as much to the table .\end{tabular} \\ \hline
DAN  & \begin{tabular}[c]{@{}l@{}}fails help bring as much to the coming . \end{tabular}     \\ \hdashline
CNN  & \begin{tabular}[c]{@{}l@{}}take to bring it this to at you  \textless{}u\textgreater{} . a . makes , so \end{tabular}                  \\ \hdashline
RNN  & \begin{tabular}[c]{@{}l@{}}\textbf{manages} to bring as much to the table .\end{tabular}                        \\ \toprule

Orig & \begin{tabular}[c]{@{}l@{}} now it 's a \textbf{bad} , \textbf{embarrassing} movie .\end{tabular} \\ \hline
DAN  & \begin{tabular}[c]{@{}l@{}}now it 's a bad , embarrassing movie . \end{tabular}     \\ \hdashline
CNN  & \begin{tabular}[c]{@{}l@{}}now it 's a good , enough movie . this it makes it makes , makes , makes  \textless{}u\textgreater{} . a . makes , so \end{tabular}                  \\ \hdashline
RNN  & \begin{tabular}[c]{@{}l@{}}now it 's a \textbf{good} , \textbf{unexpected} movie .\end{tabular}                        \\ \toprule

Orig & \begin{tabular}[c]{@{}l@{}} an often - deadly \textbf{boring} , strange reading of a classic whose witty dialogue is treated with a \\ \textless{}u\textgreater{} casual approach\end{tabular} \\ \hline
DAN  & \begin{tabular}[c]{@{}l@{}}an often - full boring , strange reading of a theme whose witty dialogue is treated with a \\ \textless{}u\textgreater{} casual yet \end{tabular}     \\ \hdashline
CNN  & \begin{tabular}[c]{@{}l@{}}an often - a pleasant , strange reading of a classic whose charming dialogue is taking with a \\ . life  \end{tabular}                  \\ \hdashline
RNN  & \begin{tabular}[c]{@{}l@{}}an often - deadly \textbf{hilarious} , strange reading of a classic whose witty dialogue is treated with \\ a \textless{}u\textgreater{} perspective come \end{tabular}                        \\ \toprule

Orig & \begin{tabular}[c]{@{}l@{}} a gimmick in search of a movie : how to get \textless{}u\textgreater{} into as many \textbf{silly} costumes and deliver as \\ many silly voices as possible , plot mechanics be damned .\end{tabular} \\ \hline
DAN  & \begin{tabular}[c]{@{}l@{}}a gimmick in search of a theme : see to get \textless{}u\textgreater{} into as many \textbf{good} costumes and deliver as \\ many good voices as possible , plot mechanics be damned while \end{tabular}     \\ \hdashline
CNN  & \begin{tabular}[c]{@{}l@{}}a offers in search of a film : how to take \textless{}u\textgreater{} up also many fun costumes and deliver also \\ many good and as this , one skills be this this on to a . makes often  \end{tabular}                  \\ \hdashline
RNN  & \begin{tabular}[c]{@{}l@{}}a riveting in connection of a movie : how to get \textless{}u\textgreater{} into as many silly costumes and deliver \\ as many silly voices as possible , plot mechanics be damned come \end{tabular}                        \\ \toprule

\end{tabular}
\caption{Example output for SIFT in the \textbf{Artifact Detection} setting of the different classifiers compared to the original on SST-2. In this case we have flipped the labels from \textbf{negative to positive}. }
\label{table:examples_sentiment_1}\end{table*}

\renewcommand{\arraystretch}{1.1}
\begin{table*}[htp]
\centering
\begin{tabular}{ll}
\toprule

Orig & \begin{tabular}[c]{@{}l@{}} invincible is a wonderful movie .\end{tabular} \\ \hline
DAN  & \begin{tabular}[c]{@{}l@{}}invincible is a wonderful movie however \end{tabular}     \\ \hdashline
CNN  & \begin{tabular}[c]{@{}l@{}}this is a little movie  \end{tabular}                  \\ \hdashline
RNN  & \begin{tabular}[c]{@{}l@{}} wes is a pathetic movie .\end{tabular}                        \\ \toprule

Orig & \begin{tabular}[c]{@{}l@{}} sharp , lively , funny and ultimately sobering film .\end{tabular} \\ \hline
DAN  & \begin{tabular}[c]{@{}l@{}}sharp , lively , funny and ultimately sobering film .\end{tabular}     \\ \hdashline
CNN  & \begin{tabular}[c]{@{}l@{}}and , little , little and which this film  \end{tabular}                  \\ \hdashline
RNN  & \begin{tabular}[c]{@{}l@{}} sharp , awkward , joke and ultimately puzzling film .\end{tabular}                        \\ \toprule

Orig & \begin{tabular}[c]{@{}l@{}} an exciting and involving rock music doc , a smart and satisfying look inside that \\ tumultuous world . \end{tabular} \\ \hline
DAN  & \begin{tabular}[c]{@{}l@{}}an interesting and involving rock music doc , another wise and satisfying things inside that \\ tumultuous series however \end{tabular}     \\ \hdashline
CNN  & \begin{tabular}[c]{@{}l@{}}an course and this rock music doc , a little and kind even inside that half this . \end{tabular}                  \\ \hdashline
RNN  & \begin{tabular}[c]{@{}l@{}} an boring and involving rock music doc , a smart and predictable looking inside that \\ dilapidated world .\end{tabular}                        \\ \toprule

\end{tabular}
\caption{Example output for SIFT in the \textbf{Artifact Detection} setting of the different classifiers compared to the original on SST-2. In this case we have flipped the labels from \textbf{positive to negative}. }
\label{table:examples_sentiment_2}\end{table*}


\renewcommand{\arraystretch}{1.1}
\begin{table*}[htp]
\centering
\begin{tabular}{ll}
\toprule

Orig & \begin{tabular}[c]{@{}l@{}}rates an ` e ' for effort -- and a ` b ' for boring .\end{tabular} \\ \hline
DAN  & \begin{tabular}[c]{@{}l@{}}rates ` e e effort effort -- -- -- ` b b b boring boring boring boring boring\end{tabular}     \\ \hdashline
CNN  & \begin{tabular}[c]{@{}l@{}}rates on ` bad on on on in in in ' an ` ' a boring '' \end{tabular}                  \\ \hdashline
RNN  & \begin{tabular}[c]{@{}l@{}}rates called ` 8 \textless{}u\textgreater{} from \textless{}u\textgreater{} while , of ` b b of boring boring \end{tabular}                        \\ \toprule

Orig & \begin{tabular}[c]{@{}l@{}}if your senses have n't been \textless{}u\textgreater{} by \textless{}u\textgreater{} films and \textless{}u\textgreater{} , if you 're  \\ a \textless{}u\textgreater{} of  psychological horror ,  this is your ticket .\end{tabular} \\ \hline
DAN  & \begin{tabular}[c]{@{}l@{}}if senses senses have n't n't been \textless{}u\textgreater{} \textless{}u\textgreater{} \textless{}u\textgreater{} films films films , \textless{}u\textgreater{} , you 're  \\ psychological psychological horror this is your ticket ticket ticket ticket \end{tabular}     \\  \hdashline
CNN  & \begin{tabular}[c]{@{}l@{}}but your how hard out on in in in ( ( ( ( ( ( ( on about about about about horror this this \\ your ticket '' \end{tabular}                  \\  \hdashline
RNN  & \begin{tabular}[c]{@{}l@{}}it your desire \textless{}u\textgreater{} n't been \textless{}u\textgreater{} by \textless{}u\textgreater{} \textless{}u\textgreater{} , \textless{}u\textgreater{} , \textless{}u\textgreater{} you you \textless{}u\textgreater{} \textless{}u\textgreater{} of \\ psychological  horror , this is your bet bet limit limit \end{tabular}                        \\ \bottomrule

Orig & \begin{tabular}[c]{@{}l@{}}\textless{}u\textgreater{} turns in a \textless{}u\textgreater{} screenplay that \textless{}u\textgreater{} at the edges ; it 's so clever you want to hate it .\end{tabular} \\ \hline
DAN  & \begin{tabular}[c]{@{}l@{}}\textless{}u\textgreater{} turns in a \textless{}u\textgreater{} screenplay screenplay screenplay of \textless{}u\textgreater{} edges edges edges shapes \\ so clever  easy want hate hate hate hate hate hate hate hate hate hate \end{tabular}     \\ \hdashline
CNN  & \begin{tabular}[c]{@{}l@{}}she turns on a on ( ( in in the the the edges 's so clever `` want to hate it '' \end{tabular}                  \\ \hdashline
RNN  & \begin{tabular}[c]{@{}l@{}}\textless{}u\textgreater{} turns in a \textless{}u\textgreater{} screenplay was \textless{}u\textgreater{} \textless{}u\textgreater{} \textless{}u\textgreater{} edges edges edges curves \textless{}u\textgreater{} clever \\ clever  you want hate hate it . \end{tabular}                        \\ \bottomrule

Orig & \begin{tabular}[c]{@{}l@{}}a pleasant \textless{}u\textgreater{} through the sort of \textless{}u\textgreater{} terrain that \textless{}u\textgreater{} morris has often dealt with ... it \\ does possess a loose , \textless{}u\textgreater{} charm .\end{tabular} \\ \hline
DAN  & \begin{tabular}[c]{@{}l@{}}a pleasant \textless{}u\textgreater{} through the sort kind \textless{}u\textgreater{} terrain terrain terrain terrain \textless{}u\textgreater{} morris \\ mitchell dealt dealt  ... does possess loose loose voice wit charm charm charm charm \end{tabular}     \\ \hdashline
CNN  & \begin{tabular}[c]{@{}l@{}}a pleasant \textless{}u\textgreater{} through on on in in ( ( ( ( ( ( ( on on that that about about about a \textless{}u\textgreater{} \\ charm ''  \end{tabular}                  \\ \hdashline
RNN  & \begin{tabular}[c]{@{}l@{}}a pleasant \textless{}u\textgreater{} through the idea of \textless{}u\textgreater{} woven of \textless{}u\textgreater{} \textless{}u\textgreater{} \textless{}u\textgreater{} often \textless{}u\textgreater{} \textless{}u\textgreater{} \\ \textless{}u\textgreater{}  which may  can a loose , \textless{}u\textgreater{} wit charm \end{tabular}                        \\ \bottomrule

\end{tabular}
\caption{Example output for SIFT in the Classifier Inspection setting of the different classifiers compared to the original on SST-2. We can see that DAN keeps repeating words that have high sentiment value. CNN masks out words with `(' and stop-words (`about', `on', etc.). RNN uses \textless{}u\textgreater{} or repeats previous words as masking operators.}
\label{table:examples_sentiment_3}\end{table*}

\renewcommand{\arraystretch}{1.1}
\begin{table*}[htp]
\centering
\begin{tabular}{ll}
\toprule

Orig & \begin{tabular}[c]{@{}l@{}} there was no significant difference in overall survival between groups ( median overall \\ survival 128 months [ \textless{}u\textgreater{} \% ci 105 - 143 ] in the \textless{}u\textgreater{} group vs 143 months [ 125 - 165 ] \\ in the \textless{}u\textgreater{} and \textless{}u\textgreater{} group ; hazard ratio [ hr ] \textless{}u\textgreater{} [ \textless{}u\textgreater{} \% ci \textless{}u\textgreater{} - 103 ] ; \textless{}u\textgreater{} log - \\ rank test p = \textless{}u\textgreater{} ) . \end{tabular} \\ \hline
DAN  & \begin{tabular}[c]{@{}l@{}}there was no significant difference by overall  survival between groups ( median overall \\ individual [ [ [ \textless{}u\textgreater{} \% ci 105 - 143 ] in the \textless{}u\textgreater{}  group vs 143 months [ 125 - 165 ] in the \\ \textless{}u\textgreater{} and \textless{}u\textgreater{} group ; hazard ratio [ hr ] \textless{}u\textgreater{} [ \textless{}u\textgreater{} \% ci \textless{}u\textgreater{} - 103 ] ; \textless{}u\textgreater{} log - rank \\ test p = \textless{}u\textgreater{} ) . \end{tabular}     \\ \hdashline
CNN  & \begin{tabular}[c]{@{}l@{}} there was no significant difference in overall survival between groups : median overall [ = \\ - - - , , ] , , ] \textless{}u\textgreater{} \textless{}u\textgreater{} group group , 161 months [ 125 - 165 ] in the \textless{}u\textgreater{} and \textless{}u\textgreater{} group ; \\ hazard ratio [ hr ] \textless{}u\textgreater{} [ \textless{}u\textgreater{} \% ci \textless{}u\textgreater{} - 103 ] ; \textless{}u\textgreater{}  log - rank test p = \textless{}u\textgreater{} ) . \end{tabular}                  \\ \hdashline
RNN  & \begin{tabular}[c]{@{}l@{}}there was no significant difference in overall survival \textless{}u\textgreater{} \textless{}u\textgreater{} - \textless{}u\textgreater{} \textless{}u\textgreater{} \textless{}u\textgreater{} \textless{}u\textgreater{} vs \\ \textless{}u\textgreater{} - \textless{}u\textgreater{} ] in the \textless{}u\textgreater{} group vs \textless{}u\textgreater{} \textless{}u\textgreater{} [ 125 - 165 ] in the \textless{}u\textgreater{} and \textless{}u\textgreater{} group ; \\ \textless{}u\textgreater{} group \textless{}u\textgreater{} hazard ratio hr ] \textless{}u\textgreater{} [ \textless{}u\textgreater{} \textless{}u\textgreater{} ci \textless{}u\textgreater{} ci \textless{}u\textgreater{} \textless{}u\textgreater{} - \textless{}u\textgreater{} ] ; \textless{}u\textgreater{} \\ log - rank p = \textless{}u\textgreater{} \end{tabular}                        \\ \toprule

Orig & \begin{tabular}[c]{@{}l@{}} study 1 : under \textless{}u\textgreater{} conditions , a separation between \textless{}u\textgreater{} and placebo on minute \\ ventilation was observed by 6.1 ( 3.6 to 8.6 ) l / min ( p \textless{} 0.01 ) and 3.6 ( 1.5 to 5.7 ) l / min \\ ( p \textless{} 0.01 ) at low - dose \textless{}u\textgreater{} plus high - dose \textless{}u\textgreater{} and high - dose - \textless{}u\textgreater{} plus high - dose \end{tabular} \\ \hline
DAN  & \begin{tabular}[c]{@{}l@{}}study 1 or by \textless{}u\textgreater{} conditions , a separation between \textless{}u\textgreater{} and placebo on minute \\ ventilation was observed by mis ( 3.6 to 8.6 ) l / min ( p \textless{} 0.01 ) and 3.6  ( 1.5 to 5.7 ) l / min \\ ( p \textless{} 0.01 ) at low - dose \textless{}u\textgreater{} plus  high - dose \textless{}u\textgreater{} and high - dose - \textless{}u\textgreater{} plus  high - dose \end{tabular}     \\ \hdashline
CNN  & \begin{tabular}[c]{@{}l@{}} study 1 : under \textless{}u\textgreater{} conditions , a separation between \textless{}u\textgreater{} and placebo at 8.2 , : 3.6 , \\ 3.6 , 3.4 ) = , l , min , p \textless{} 0.01 ) and 3.6 ( 1.5 to 6.3 ) l / min ( p \textless{} 0.01 ) at low - dose \textless{}u\textgreater{} \\ plus high - dose \textless{}u\textgreater{} and high - dose - \textless{}u\textgreater{} plus high - dose  \end{tabular}                  \\ \hdashline
RNN  & \begin{tabular}[c]{@{}l@{}}study 1 : by \textless{}u\textgreater{} conditions , a \textless{}u\textgreater{} and \textless{}u\textgreater{} and \textless{}u\textgreater{} - \textless{}u\textgreater{} , , \textless{}u\textgreater{} \textless{}u\textgreater{} min / min , \\ min , \textless{}u\textgreater{} min , p \textless{} 0.01 , and 3.6 - 1.5 - \textless{}u\textgreater{} \textless{}u\textgreater{} min / min ( p \textless{} 0.01 ) at low - dose \\ \textless{}u\textgreater{} plus high - \textless{}u\textgreater{} \textless{}u\textgreater{} and high - dose - \textless{}u\textgreater{} plus \textless{}u\textgreater{} - \end{tabular}                        \\ \toprule

\end{tabular}
\caption{Example output for SIFT in the Classifier Inspection setting of the different classifiers compared to the original on PubMed. Similar to SST-2 we can see that CNN masks out words with punctuation and RNN uses \textless{}u\textgreater{}.}
\label{table:examples_PubMed}\end{table*}


\end{document}

%% file: 05-analysis.tex
\section{Experimental Setup} \label{sec:experimental-setup}

We perform experiments on three sentence classification tasks:
1) Sentiment analysis 
\cite[SST-2;][]{Socher2013};
2) Natural language inference, 
\cite[SNLI;][]{Bowman2015}; and
3) PubMed classification 
\cite[PubMed;][]{Dernoncourt2017}.
We ensure that the pre-trained AE achieves $99\%$ token-level accuracy on the respective data sets establishing that it is able to adequately reproduce the input in the initial pretraining phase. 

\begin{table}[htp]
\centering
\small{
\begin{tabular}{ll}
\toprule
Orig & \begin{tabular}[c]{@{}l@{}}\textless{}u\textgreater{} turns in a \textless{}u\textgreater{} screenplay that \textless{}u\textgreater{} at the \\ edges ; it 's so clever you want to hate it .\end{tabular} \\ \hline
DAN  & \begin{tabular}[c]{@{}l@{}}\textless{}u\textgreater{} turns in a \textless{}u\textgreater{} screenplay screenplay \\ screenplay of \textless{}u\textgreater{} edges edges edges shapes  so \\ clever  easy want hate  hate hate hate hate hate \\ hate  hate hate hate \end{tabular}     \\ \hdashline
CNN  & \begin{tabular}[c]{@{}l@{}}she turns on a on ( ( in in the the the edges 's so \\ clever `` want to hate it '' \end{tabular}                  \\ \hdashline
RNN  & \begin{tabular}[c]{@{}l@{}}\textless{}u\textgreater{} turns in a \textless{}u\textgreater{} screenplay was \textless{}u\textgreater{} \textless{}u\textgreater{} \\ \textless{}u\textgreater{} edges edges edges curves \textless{}u\textgreater{} clever \\  clever  you want hate  hate it . \end{tabular}                        \\ \bottomrule

\end{tabular}
}
\caption{Example sentences of the different classifiers compared to the original on SST-2. We report further examples in the Appendix. \textless{}u\textgreater{} use
for \textless{}UNK\textgreater{}.}
\label{table:examples_sentiment_inside}\end{table}

\subsection{SIFT for Classifier Inspection} \label{sec:classifier-comparison}


In contrast to \citet{Lei2016} who \textit{extract} rationales by enforcing a constraint that actively reduces the number of words, we do not impose any prior in an \textit{abstractive} attempt to understand the model's preferences.
Using SIFT we observed substantial differences in the text generated by our fine-tuned decoder when trained with each of the different models. 
In order to identify differences, we conducted an automated study of the reconstructed text by inspecting changes in the proportion of part-of-speech tags\footnote{NLTK \cite{loper2002nltk} was used for POS tagging.}
and an increase or decrease in word polarity for sentiment compared to the original input. Examples are in Table \ref{table:examples_sentiment_inside}, results in  \ref{table:POS_changes_SST} and \ref{table:sentiment_SST}.



\begin{table}[!htp]
\centering{\footnotesize
\begin{tabular}{llll}
\toprule
                           & \textbf{RNN}  & \textbf{CNN}  & \textbf{DAN}  \\ \toprule
Nouns                      & \tikztriangleup ~ $+63$  & \tikz\draw[gray,fill=gray] (0,0) circle (.7ex);  $-3$ & \tikztriangleup    ~ $+\textbf{93}$   \\ 
DT                         & \tikztriangledown  ~ $-29$ & \tikztriangleup ~ $+32$ & \tikztriangledown ~ $-\textbf{38}$ \\ 
Verbs                      & \tikztriangleup ~ $+20$ & \tikz\draw[gray,fill=gray] (0,0) circle (.7ex);  $-4$ & \tikztriangleup ~ $+\textbf{34}$  \\ 
Adj.                       & \tikztriangleup ~  $+25$ & \tikz\draw[gray,fill=gray] (0,0) circle (.7ex);  $-1$ & \tikztriangleup ~ $+\textbf{66}$ \\ 
Prep.                      & \tikztriangleup ~ $+12$ & \tikztriangleup ~ $+12$ & \tikztriangledown ~ $-\textbf{62}$ \\ 
Punct.                     & \tikztriangledown ~ $-\textbf{53}$ & \tikztriangledown ~ $-14$ & \tikztriangledown ~ $-47$ \\ 
\textless{}U\textgreater{} & \tikztriangleup ~ $+\textbf{82}$   & \tikztriangledown ~ $-14$ & \tikztriangleup ~ $+16$ \\
\textbf{RNP} & $69.0\%$   & $70.5\%$ & $81.5\%$ \\
\bottomrule
\end{tabular}\caption{Part-of-Speech (POS) changes in SST-2: \tikztriangleup, \tikztriangledown , and \tikz\draw[gray,fill=gray] (0,0) circle (.7ex);  indicate that the number of occurrences have increased, decreased or stayed the same through fine-tuning respectively. The symbols are purely analytic without any notion of \textit{goodness}. The numbers indicate the changes in percentage points with respect to the original sentence. A score of 0 thus means that fine-tuning has not changed the number of words. The last row indicates the overlap with the extractive RNP approach. We report results for PubMed in the Appendix.}
\label{table:POS_changes_SST}}

\end{table}

\begin{table}[ht]
\centering{\footnotesize
\begin{tabular}{llll}
\toprule
                           & \textbf{RNN}  & \textbf{CNN}  & \textbf{DAN}  \\ \toprule
Positive                      & \tikztriangleup ~ +$9.7$   & \tikztriangleup+$4.3$   & \tikztriangleup +$\boldsymbol{23.6}$     \\ 
Negative    & \tikztriangledown ~ +$6.9$   & \tikztriangledown +$5.5$   & \tikztriangledown +$\boldsymbol{16.1}$     \\ 
Flipped to Positive                      & \tikztriangleup ~ +$20.2$   & \tikztriangleup$+24.9$   & \tikztriangleup +$27.4$     \\ 
Flipped to Negative    & \tikztriangledown ~ +$31.5$   & \tikztriangledown +$28.6$   & \tikztriangledown +$19.3$     \\ 

\bottomrule
\end{tabular}\caption{Sentiment score changes in SST-2. The numbers indicate the changes in percentage points with respect to the original sentence. The last two rows correspond to the case where negative labels are flipped to positive and vice versa. \tikztriangleup ~ and \tikztriangledown ~ indicate that the score increases in positive and negative sentiment.}
\label{table:sentiment_SST}}
\end{table}

Similar to the extractive approach of \citet{Lei2016}, who \textit{actively} mask out terms by extracting them, we find that all three classifiers \textit{implicitly} mask out words. While the RNN primarily employs $<$UNK$>$ tokens or repeats previous words, the CNN \textit{masks out} $<$UNK$>$ tokens using determiners or prepositions. In contrast, DAN masks out punctuation and determiners using words indicative of the class label (i.e. nouns, verbs, adjectives). We hypothesize that these patterns stem from the inductive biases of the classifiers. DAN receives a stronger signal by repeating words with a higher sentiment value due to its averaging, while the CNN does not repeat words (thus having the least amount of changes) and removes uninformative words as its max-pooling layer selects only the most important ones.
Similarly, the gates of the LSTM may allow the model to ignore the random and thus noisy $<$UNK$>$ embeddings, which enables it to use this token as a masking operation to ignore unimportant words.

To compare our \textit{abstractive} with an \textit{extractive} approach \cite[RNP;][]{Lei2016}, we compute the overlap of retained terms in Table \ref{table:POS_changes_SST} (bottom row). We can see that the DAN has the highest overlap, indicating that it retains words, while the CNN and RNN reformulate sentences. These scores highlight the differences of our approach, as our model does not solely extract indicative words, but reformulates the original sentence. 

In order to automatically identify if SIFT retains the sentiment of the sentences, we analyze the output using SentiWordNet \cite{baccianella2010sentiwordnet}. By considering only adjectives, we obtain a measure of the positive and negative score for each sentence before and after fine-tuning. The difference of these scores averaged over all examples provides us with a sense of whether the fine-tuning increases the polarity of the sentences (Table \ref{table:sentiment_SST}). We see a constant increase in sentiment value in both directions across all three models after fine-tuning demonstrating that the framework is able to pick up on words that are indicative of sentiment. This is especially true in the case of DAN where we see a large increase as the decoder repeatedly predicts words having high sentiment value. Overall, these results indicate that SIFT is able to highlight certain inductive biases of the model and is able to reformulate and amplify the meaning of the original text based on the classifier's preferences.

\subsection{SIFT for Artifact Detection} \label{sec:opposite-labels}

Much recent work has focused on detecting artifacts, i.e. in data generated and annotated by humans, with findings that suggest supervised models heavily rely on their existence \cite{Levy2015, Gururangan2018, May2019}. While e.g. pointwise mutual information (PMI) has been used to identify terms indicative of a specific class \cite{Gururangan2018}, it is unclear if a model actually encodes these terms. 

Intuitively, such terms may be a source of bias if their presence in the input causes the classifier to change its prediction to their respective class.
To uncover these encoded artifacts we fine-tune SIFT on data where the ground truth label of all examples of a particular class are \emph{changed to that of the other class}, and this is done subsequently in the other direction. 
Since the weights of the classifier are fixed during the fine-tuning phase, our framework forces the decoder to update its parameters to produce text that causes the classifier to reverse the label for the data point. To ensure coherent sentences that are comparable to the original versions, we enforce an additional loss that encourages similarity
by penalizing a large cosine distance (cos) between the average embedding of the sentences: 
\begin{equation} \label{eq:cosine}
\text{d}(\textbf{X}, \hat{\textbf{X}}) =  \text{cos}(\frac{1}{N} \sum_{n\in N} \textbf{X}_n , \frac{1}{M} \sum_{m\in M} \hat{\textbf{X}}_m)
\end{equation}
with  $\textbf{X}$ and $\hat{\textbf{X}}$ being the word embeddings of the original and  generated sentence of lengths $N$, $M$. 


For this setting we apply SIFT to a sentence pair classification task, i.e. SNLI \cite{Bowman2015} and examine the case where we flip the labels of data points from \textit{entailment} to \textit{contradiction}. 

\begin{table}[]
\centering{\footnotesize
\begin{tabular}{ c ll ll }
\toprule
\multicolumn{1}{l}{} & \multicolumn{2}{c}{\textbf{SST-2}} & \multicolumn{2}{c}{\textbf{PubMed}}  \\
\multicolumn{1}{l}{} & Positive         & Negative        & Objective          & Conclusion      \\ \cmidrule(ll){2-3} \cmidrule(ll){4-5} 
\multirow{3}{*}{\textbf{PMI}} & best           & too          & compare            & should                     \\
                      & love           &  bad         & investigate            & suggest                 \\
                      & fun           &  n't         & evaluate            & findings                
                    \\ \cmidrule(ll){2-3} \cmidrule(ll){4-5} 
\multirow{3}{*}{\textbf{SIFT}}  & but           & nothing          & to            & larger                  \\
                      & come           & awkward          & whether            & confirm                 \\
                      & it           & lacking          & clarify            & concluded                  

                      \\ 
                        \hdashline
Acc &       $\textit{98\%}$           & $\textit{98\%}$          & $\textit{98\%}$            & $\textit{99\%}$

                      \\ \hline
 \textit{Corr}       & $\textit{0.486}$           & $\textit{0.5415}$          & $\textit{0.398}$            & $\textit{0.00089}$               
                      \\                
                      \bottomrule
\end{tabular}}
\caption{Top 3 PMI and SIFT terms for a subset of classes with SIFT on the RNN model. The second last row depicts the SIFT accuracy on the test set for the flipped label setting. The last row indicates the correlation of the PMI and SIFT list using weighted Kendall's tau correlation \cite{Shieh1998} .  }

\label{table:Premise_analysis_contradiction}
\end{table}

We compare the most frequent SIFT terms with the most indicative words w.r.t. a class using PMI with 100 smoothing, following \citet{Gururangan2018}. We list the most frequently used words in Table \ref{table:Premise_analysis_contradiction}.  To understand how similar the SIFT terms are to the PMI terms we calculate the weighted Kendall's tau correlation \cite{Shieh1998}, which weights terms higher in the list as more important.
In the case of high correlation, we hypothesize that the classifier has memorized the artifacts of the data set, which in turn SIFT has leveraged to \textit{``fool"} the classifier. This \textit{``fooling"} in the case of sentiment analysis is due to the terms being indicative of the respective class and having high sentiment (cf. Table \ref{table:sentiment_SST}). For SNLI, we find that many of the top PMI and SIFT terms overlap, and slightly correlate ($0.366$). We  find  terms (as reported in Appendix \ref{table:A2}), e.g. `sleeping', `cats', `cat', that were  identified as artifacts by \citet{Gururangan2018}.

In the PubMed task where the aim is to classify sentences as belonging to one of five classes---background, objective, methods, results, conclusions---the top terms in the setting where we change ground truth from ``objective" to ``conclusion" are intuitively relevant terms for that class. They do not, however, correlate with PMI, which might indicate that pretraining on large unlabelled data has enabled SIFT to capture these relations. 

Overall, this shows that SIFT is able to identify both previously known as well as novel artifacts. In contrast to PMI, SIFT uncovers the propensities of the trained model and not only the data set, giving insight into what the model has actually encoded. 

%% file: emnlp-ijcnlp-2019.bbl
\begin{thebibliography}{36}
\expandafter\ifx\csname natexlab\endcsname\relax\def\natexlab#1{#1}\fi

\bibitem[{Adi et~al.(2017)Adi, Kermany, Belinkov, Lavi, and Goldberg}]{Adi2017}
Yossi Adi, Einat Kermany, Yonatan Belinkov, Ofer Lavi, and Yoav Goldberg. 2017.
\newblock \href {http://arxiv.org/abs/arXiv:1608.04207v3} {{Fine-grained
  Analysis of Sentence Embeddings Using Auxiliary Prediction Tasks}}.
\newblock In \emph{Proceedings of ICLR 2017}.

\bibitem[{Baccianella et~al.(2010)Baccianella, Esuli, and
  Sebastiani}]{baccianella2010sentiwordnet}
Stefano Baccianella, Andrea Esuli, and Fabrizio Sebastiani. 2010.
\newblock Sentiwordnet 3.0: an enhanced lexical resource for sentiment analysis
  and opinion mining.
\newblock In \emph{Lrec}, volume~10, pages 2200--2204.

\bibitem[{Bahdanau et~al.(2015)Bahdanau, Cho, and Bengio}]{Bahdanau2015}
Dzmitry Bahdanau, Kyunghyun Cho, and Yoshua Bengio. 2015.
\newblock {Neural Machine Translation by Jointly Learning to Align and
  Translate}.
\newblock In \emph{Proceedings of ICLR 2015}.

\bibitem[{Bao et~al.(2018)Bao, Chang, Yu, Barzilay, and Science}]{Bao2018}
Yujia Bao, Shiyu Chang, Mo~Yu, Regina Barzilay, and Computer Science. 2018.
\newblock \href {http://arxiv.org/abs/arXiv:1808.09367v1} {{Deriving Machine
  Attention from Human Rationales}}.
\newblock In \emph{Proceedings of EMNLP 2018}.

\bibitem[{Bastings et~al.(2019)Bastings, Aziz, and Titov}]{BastingsAT19}
Joost Bastings, Wilker Aziz, and Ivan Titov. 2019.
\newblock \href {https://www.aclweb.org/anthology/P19-1284/} {Interpretable
  neural predictions with differentiable binary variables}.
\newblock In \emph{Proceedings of the 57th Conference of the Association for
  Computational Linguistics, {ACL} 2019, Florence, Italy, July 28- August 2,
  2019, Volume 1: Long Papers}, pages 2963--2977.

\bibitem[{Bau et~al.(2019)Bau, Durrani, Belinkov, Dalvi, Sajjad, and
  Glass}]{Bau2019}
Anthony Bau, Nadir Durrani, Yonatan Belinkov, Fahim Dalvi, Hassan Sajjad, and
  James Glass. 2019.
\newblock \href {http://arxiv.org/abs/arXiv:1811.01157v1} {{Identifying and
  Controlling Important Neurons in Neural Machine Translation}}.
\newblock In \emph{Proceedings of ICLR 2019}.

\bibitem[{Belinkov and Glass(2019)}]{Belinkov2019}
Yonatan Belinkov and James Glass. 2019.
\newblock \href {http://arxiv.org/abs/arXiv:1812.08951v1} {{Analysis Methods in
  Neural Language Processing: A Survey}}.
\newblock \emph{Transactions of the ACL}.

\bibitem[{Bowman et~al.(2015)Bowman, Angeli, Potts, and Manning}]{Bowman2015}
Samuel~R. Bowman, Gabor Angeli, Christopher Potts, and Christopher~D. Manning.
  2015.
\newblock A large annotated corpus for learning natural language inference.
\newblock In \emph{Proceedings of the 2015 Conference on Empirical Methods in
  Natural Language Processing (EMNLP)}. Association for Computational
  Linguistics.

\bibitem[{Buck et~al.(2018)Buck, Bulian, Ciaramita, Gajewski, Gesmundo,
  Houlsby, and Wang}]{Buck2018}
Christian Buck, Jannis Bulian, Massimiliano Ciaramita, Wojciech Gajewski,
  Andrea Gesmundo, Neil Houlsby, and Wei Wang. 2018.
\newblock {Ask the Right Questions: Active Question Reformulation with
  Reinforcement Learning}.
\newblock In \emph{Proceedings of ICLR 2018}.

\bibitem[{Chelba et~al.(2013)Chelba, Mikolov, Schuster, Ge, Brants, Koehn, and
  Robinson}]{chelba2013one}
Ciprian Chelba, Tomas Mikolov, Mike Schuster, Qi~Ge, Thorsten Brants, Phillipp
  Koehn, and Tony Robinson. 2013.
\newblock One billion word benchmark for measuring progress in statistical
  language modeling.
\newblock \emph{arXiv preprint arXiv:1312.3005}.

\bibitem[{Conneau et~al.(2018)Conneau, Kruszewski, Lample, Barrault, and
  Baroni}]{Conneau2018a}
Alexis Conneau, German Kruszewski, Guillaume Lample, Lo{\"{i}}c Barrault, and
  Marco Baroni. 2018.
\newblock \href {http://arxiv.org/abs/1805.01070} {{What you can cram into a
  single vector: Probing sentence embeddings for linguistic properties}}.
\newblock In \emph{Proceedings of ACL 2018}.

\bibitem[{Dernoncourt and Lee(2017)}]{Dernoncourt2017}
Franck Dernoncourt and Ji~Young Lee. 2017.
\newblock Pubmed 200k rct: a dataset for sequential sentence classification in
  medical abstracts.
\newblock \emph{arXiv preprint arXiv:1710.06071}.

\bibitem[{Gururangan et~al.(2018)Gururangan, Swayamdipta, Levy, Schwartz,
  Bowman, and Smith}]{Gururangan2018}
Suchin Gururangan, Swabha Swayamdipta, Omer Levy, Roy Schwartz, Samuel~R.
  Bowman, and Noah~A. Smith. 2018.
\newblock \href {https://aclanthology.info/papers/N18-2017/n18-2017}
  {Annotation artifacts in natural language inference data}.
\newblock In \emph{Proceedings of the 2018 Conference of the North American
  Chapter of the Association for Computational Linguistics: Human Language
  Technologies, NAACL-HLT, New Orleans, Louisiana, USA, June 1-6, 2018, Volume
  2 (Short Papers)}, pages 107--112.

\bibitem[{Iyyer et~al.(2015)Iyyer, Manjunatha, Boyd{-}Graber, and
  III}]{Iyyer2015}
Mohit Iyyer, Varun Manjunatha, Jordan~L. Boyd{-}Graber, and Hal~Daum{\'{e}}
  III. 2015.
\newblock \href {http://aclweb.org/anthology/P/P15/P15-1162.pdf} {Deep
  unordered composition rivals syntactic methods for text classification}.
\newblock In \emph{Proceedings of the 53rd Annual Meeting of the Association
  for Computational Linguistics and the 7th International Joint Conference on
  Natural Language Processing of the Asian Federation of Natural Language
  Processing, {ACL} 2015, July 26-31, 2015, Beijing, China, Volume 1: Long
  Papers}, pages 1681--1691.

\bibitem[{Jang et~al.(2017)Jang, Gu, and Poole}]{Jang2017}
Eric Jang, Shixiang Gu, and Ben Poole. 2017.
\newblock \href {http://arxiv.org/abs/1611.01144} {{Categorical
  Reparameterization with Gumbel-Softmax}}.
\newblock In \emph{Proceedings of ICLR 2017}.

\bibitem[{Kim(2014)}]{Kim2014}
Yoon Kim. 2014.
\newblock \href {http://aclweb.org/anthology/D/D14/D14-1181.pdf} {Convolutional
  neural networks for sentence classification}.
\newblock In \emph{Proceedings of the 2014 Conference on Empirical Methods in
  Natural Language Processing, {EMNLP} 2014, October 25-29, 2014, Doha, Qatar,
  {A} meeting of SIGDAT, a Special Interest Group of the {ACL}}, pages
  1746--1751.

\bibitem[{Lei et~al.(2016)Lei, Barzilay, and Jaakkola}]{Lei2016}
Tao Lei, Regina Barzilay, and Tommi Jaakkola. 2016.
\newblock \href {http://arxiv.org/abs/1606.04155} {{Rationalizing Neural
  Predictions}}.
\newblock In \emph{Proceedings of EMNLP 2016}.

\bibitem[{Levy et~al.(2015)Levy, Remus, Biemann, and Dagan}]{Levy2015}
Omer Levy, Steffen Remus, Chris Biemann, and Ido Dagan. 2015.
\newblock Do supervised distributional methods really learn lexical inference
  relations?
\newblock In \emph{Proceedings of the 2015 Conference of the North American
  Chapter of the Association for Computational Linguistics: Human Language
  Technologies}, pages 970--976.

\bibitem[{Li et~al.(2016)Li, Chen, Hovy, and Jurafsky}]{LiCHJ16}
Jiwei Li, Xinlei Chen, Eduard~H. Hovy, and Dan Jurafsky. 2016.
\newblock \href {http://aclweb.org/anthology/N/N16/N16-1082.pdf} {Visualizing
  and understanding neural models in {NLP}}.
\newblock In \emph{{NAACL} {HLT} 2016, The 2016 Conference of the North
  American Chapter of the Association for Computational Linguistics: Human
  Language Technologies, San Diego California, USA, June 12-17, 2016}, pages
  681--691.

\bibitem[{Loper and Bird(2002)}]{loper2002nltk}
Edward Loper and Steven Bird. 2002.
\newblock Nltk: the natural language toolkit.
\newblock \emph{arXiv preprint cs/0205028}.

\bibitem[{Maas et~al.(2011)Maas, Daly, Pham, Huang, Ng, and Potts}]{Maas2011}
Andrew~L. Maas, Raymond~E. Daly, Peter~T. Pham, Dan Huang, Andrew~Y. Ng, and
  Christopher Potts. 2011.
\newblock \href {http://www.aclweb.org/anthology/P11-1015} {Learning word
  vectors for sentiment analysis}.
\newblock In \emph{The 49th Annual Meeting of the Association for Computational
  Linguistics: Human Language Technologies, Proceedings of the Conference,
  19-24 June, 2011, Portland, Oregon, {USA}}, pages 142--150.

\bibitem[{May et~al.(2019)May, Wang, Bordia, Bowman, and Rudinger}]{May2019}
Chandler May, Alex Wang, Shikha Bordia, Samuel~R. Bowman, and Rachel Rudinger.
  2019.
\newblock \href {http://arxiv.org/abs/1903.10561} {On measuring social biases
  in sentence encoders}.
\newblock \emph{CoRR}, abs/1903.10561.

\bibitem[{Murdoch et~al.(2018)Murdoch, Liu, and Yu}]{Murdoch2018a}
W~James Murdoch, Peter~J Liu, and Bin Yu. 2018.
\newblock \href {http://arxiv.org/abs/arXiv:1801.05453v1} {{Beyond Word
  Importance: Contextual Decomposition to Extract Interactions from LSTMs}}.
\newblock In \emph{Proceedings of ICLR 2018}.

\bibitem[{Palacio et~al.(2018)Palacio, Folz, Hees, Raue, Borth, and
  Dengel}]{Palacio2018}
Sebastian Palacio, Joachim Folz, J{\"{o}}rn Hees, Federico Raue, Damian Borth,
  and Andreas Dengel. 2018.
\newblock \href {https://doi.org/10.1109/CVPR.2018.00328} {{What do Deep
  Networks Like to See?}}
\newblock In \emph{Proceedings of CVPR 2018}.

\bibitem[{Pennington et~al.(2014)Pennington, Socher, and
  Manning}]{Pennington2014}
Jeffrey Pennington, Richard Socher, and Christopher~D. Manning. 2014.
\newblock \href {https://doi.org/10.3115/v1/D14-1162} {{Glove: Global Vectors
  for Word Representation}}.
\newblock In \emph{Proceedings of the 2014 Conference on Empirical Methods in
  Natural Language Processing}, pages 1532--1543.

\bibitem[{Peters et~al.(2018)Peters, Neumann, Zettlemoyer, Yih, Allen, and
  Science}]{Peters2018a}
Matthew~E Peters, Mark Neumann, Luke Zettlemoyer, Wen-tau Yih, Paul~G Allen,
  and Computer Science. 2018.
\newblock \href {http://arxiv.org/abs/arXiv:1808.08949v1} {{Dissecting
  Contextual Word Embeddings: Architecture and Representation}}.
\newblock In \emph{Proceedings of EMNLP 2018}.

\bibitem[{Poliak et~al.(2018)Poliak, Naradowsky, Haldar, Rudinger, and
  Van~Durme}]{poliak2018hypothesis}
Adam Poliak, Jason Naradowsky, Aparajita Haldar, Rachel Rudinger, and Benjamin
  Van~Durme. 2018.
\newblock Hypothesis only baselines in natural language inference.
\newblock In \emph{Proceedings of the Seventh Joint Conference on Lexical and
  Computational Semantics}, pages 180--191.

\bibitem[{Ribeiro et~al.(2016)Ribeiro, Singh, and Guestrin}]{Ribeiro2016}
Marco~Tulio Ribeiro, Sameer Singh, and Carlos Guestrin. 2016.
\newblock {"Why Should I Trust You ?" Explaining the Predictions of Any
  Classifier}.
\newblock In \emph{Proceedings of the 22nd ACM SIGKDD International Conference
  on Knowledge Discovery and Data Mining.} ACM.

\bibitem[{Ribeiro et~al.(2018)Ribeiro, Singh, and Guestrin}]{Ribeiro2018}
Marco~Tulio Ribeiro, Sameer Singh, and Carlos Guestrin. 2018.
\newblock {Anchors: High-Precision Model-Agnostic Explanations}.
\newblock In \emph{Proceedings of AAAI 2018}.

\bibitem[{Sakaguchi et~al.(2019)Sakaguchi, Bras, Bhagavatula, and
  Choi}]{sakaguchi2019winogrande}
Keisuke Sakaguchi, Ronan~Le Bras, Chandra Bhagavatula, and Yejin Choi. 2019.
\newblock Winogrande: An adversarial winograd schema challenge at scale.
\newblock \emph{arXiv preprint arXiv:1907.10641}.

\bibitem[{Shieh(1998)}]{Shieh1998}
Grace~S Shieh. 1998.
\newblock A weighted kendall's tau statistic.
\newblock \emph{Statistics \& probability letters}, 39(1):17--24.

\bibitem[{Socher et~al.(2013)Socher, Perelygin, Wu, Chuang, Manning, Ng, and
  Potts}]{Socher2013}
Richard Socher, Alex Perelygin, Jean Wu, Jason Chuang, Christopher~D. Manning,
  Andrew~Y. Ng, and Christopher Potts. 2013.
\newblock \href {https://aclanthology.info/papers/D13-1170/d13-1170} {Recursive
  deep models for semantic compositionality over a sentiment treebank}.
\newblock In \emph{Proceedings of the 2013 Conference on Empirical Methods in
  Natural Language Processing, {EMNLP} 2013, 18-21 October 2013, Grand Hyatt
  Seattle, Seattle, Washington, USA, {A} meeting of SIGDAT, a Special Interest
  Group of the {ACL}}, pages 1631--1642.

\bibitem[{Sutskever et~al.(2014)Sutskever, Vinyals, and Le}]{Sutskever2014}
Ilya Sutskever, Oriol Vinyals, and Quoc~V Le. 2014.
\newblock \href
  {http://papers.nips.cc/paper/5346-sequence-to-sequence-learning-with-neural-networks.pdf}
  {Sequence to sequence learning with neural networks}.
\newblock In Z.~Ghahramani, M.~Welling, C.~Cortes, N.~D. Lawrence, and K.~Q.
  Weinberger, editors, \emph{Advances in Neural Information Processing Systems
  27}, pages 3104--3112. Curran Associates, Inc.

\bibitem[{Tsuchiya(2018)}]{tsuchiya2018performance}
Masatoshi Tsuchiya. 2018.
\newblock Performance impact caused by hidden bias of training data for
  recognizing textual entailment.
\newblock In \emph{Proceedings of the Eleventh International Conference on
  Language Resources and Evaluation (LREC-2018)}.

\bibitem[{Zellers et~al.(2018)Zellers, Bisk, Schwartz, and
  Choi}]{zellers2018swag}
Rowan Zellers, Yonatan Bisk, Roy Schwartz, and Yejin Choi. 2018.
\newblock Swag: A large-scale adversarial dataset for grounded commonsense
  inference.
\newblock In \emph{Proceedings of the 2018 Conference on Empirical Methods in
  Natural Language Processing}, pages 93--104.

\bibitem[{Zhu et~al.(2018)Zhu, Li, and {De Melo}}]{Zhu2018}
Xunjie Zhu, Tingfeng Li, and Gerard {De Melo}. 2018.
\newblock \href {http://aclweb.org/anthology/P18-2100} {{Exploring Semantic
  Properties of Sentence Embeddings}}.
\newblock In \emph{Proceedings of ACL 2018}.

\end{thebibliography}
